\documentclass[10pt]{article}
\usepackage[preprint]{tmlr}

\usepackage[utf8]{inputenc}
\usepackage[T1]{fontenc}
\usepackage[hidelinks]{hyperref}
\usepackage{url}
\usepackage{booktabs}
\usepackage{amsmath}
\usepackage{amssymb}
\usepackage{graphicx}
\usepackage{array}
\usepackage{microtype}
\usepackage{xcolor}

\hypersetup{pdftitle={LLM Judges Have Dark Current: Stratified Null and Utility Tests for LLM Judge Reliability},pdfauthor={Hiroyasu Usami, Keisuke Hara, Ayato Tsuboi, Naohiko Matsuda}}

\newcommand{\DCcore}{\ensuremath{\mathrm{DC}_{\mathrm{core}}}}
\newcommand{\DCedge}{\ensuremath{\mathrm{DC}_{\mathrm{edge}}}}
\newcommand{\EFP}{\ensuremath{\mathrm{EFP}}}
\newcommand{\RFP}{\ensuremath{\mathrm{RFP}_{0}}}
\newcommand{\SCS}{\ensuremath{\mathrm{SCS}}}
\newcommand{\PFP}{\ensuremath{\mathrm{PFP}}}
\newcommand{\ind}{\mathbb{I}}

\title{LLM Judges Have Dark Current:\\Stratified Null and Utility Tests for LLM Judge Reliability\thanks{The research questions and objectives were defined by the authors. AI-assisted tools supported implementation review and refinement, manuscript organization, drafting and language revision, and internal critique. Assistance outputs do not constitute independent validation of the experimental results. The authors take full responsibility for the final manuscript.}}
\author{\name Hiroyasu Usami$^1$ \email usami@fsc.chubu.ac.jp \\
\name Keisuke Hara$^1$, Ayato Tsuboi$^1$, and Naohiko Matsuda$^2$ \\
\addr $^1$Department of Computer Science, Graduate School of Engineering, Chubu University,\\
Kasugai, Aichi 487--8501, Japan\\
\addr $^2$Mitsubishi Heavy Industries, Ltd., Research \& Innovation Center,\\
Heat Transfer Research Department, Takasago, Hyogo 676--8686, Japan}
\date{}

\begin{document}
\lhead{Preprint}
\maketitle

\begin{abstract}
Passing an exact-clone null gate does not entail passing a matched target-selection utility gate. We call a non-tie preference between exact duplicates dark current. On 102 prespecified fresh RewardBench~2 rows, all four configuration--criterion cells report 0/102 clone preferences. Both neutral utility arms pass, but strict utility lower bounds of .539 and .745 miss the .75 working threshold. Conversely, re-evaluation of the same rows on two additional configurations yields a neutral gpt-oss utility pass with 2/102 clone preferences, leaving low-null behavior uncertified. A separately frozen 485-row extension passes both aggregate working gates under both criteria on one local raw-text configuration, yet strict Precise Instruction Following (Precise IF) alignment is 58/154 (.377). Thus matched null and utility results can diverge, and aggregate passage can coexist with weak subset alignment. Interpretation is conditional on the declared configurations, fixed source mixtures, source-label targets, and study-specific working models. The bounds are marginal, the subset rate is descriptive, and non-certification does not prove a defect.
\end{abstract}

\section{Introduction}

LLM-as-a-judge systems have become evaluation infrastructure. MT-Bench and Chatbot Arena demonstrated their practical scale, while also documenting sensitivity to presentation and response attributes \citep{zheng2023mtbench}. Subsequent work has measured position bias \citep{shi2025positionbias}, multidimensional bias \citep{ye2025justice}, benchmark-level evaluator competence \citep{wang2025dhp}, and latent reliability with item-response theory (IRT) \citep{choi2026irt}. A reported score therefore characterizes a measurement pipeline, not only the evaluated model.

This distinction matters when the same judge supports scientific claims, model selection, or best-of-$N$ search. High aggregate agreement can coexist with poor decisions because the deployment distribution and error costs differ from the validation benchmark \citep{landesberg2026causal,landesberg2026bestofn}. Null models can exploit evaluator regularities to obtain high apparent win rates \citep{zheng2025null}. Disagreement is not always judge error either: some pairs admit multiple defensible ratings \citep{guerdan2025indeterminacy}. An audit must separate absence of signal, nuisance sensitivity, target sensitivity, abstention, and downstream loss.

Here, \emph{dark current} denotes a non-tie response to two exactly identical inputs. The term is an operational analogy, not a physical model of an LLM. Exact-clone behavior must also be stratified. A nonblank text duplicate probes judging on ordinary text, whereas empty and whitespace-only inputs probe pipeline edge handling. Pooling these strata can make an edge-case failure appear to be a general property of the judge.

The audit asks three questions: can the evaluation client produce eligible verdicts, does it invent a preference between exact duplicates, and does it select the declared target on signal pairs? These questions require separate evidence. The primary 102-row RewardBench~2 study supplies a null-passing, utility-missing case. Re-evaluation on additional configurations passes utility without meeting the low-null evidence criterion. The nonoverlapping 485-row extension adds a third lesson: an aggregate pass can conceal weak source-label alignment within a prespecified source subset.

The protocol combines input-stratified null tests, order-canonical response patterns, and separate operational and utility gates. Declared application losses give these measurements a decision context; their use has not been tested as a deployment intervention. Table~\ref{tab:configuration-overview} identifies the configurations across studies. Benchmark labels and finite-domain code certificates define different targets; the working statistical rules are interpreted in Section~\ref{sec:working-model}, and their scope is consolidated in Section~\ref{sec:limitations}. The purpose is to prevent unsupported adoption decisions, not to rank model families.

\begin{figure}[t]
\centering
\includegraphics[width=\linewidth]{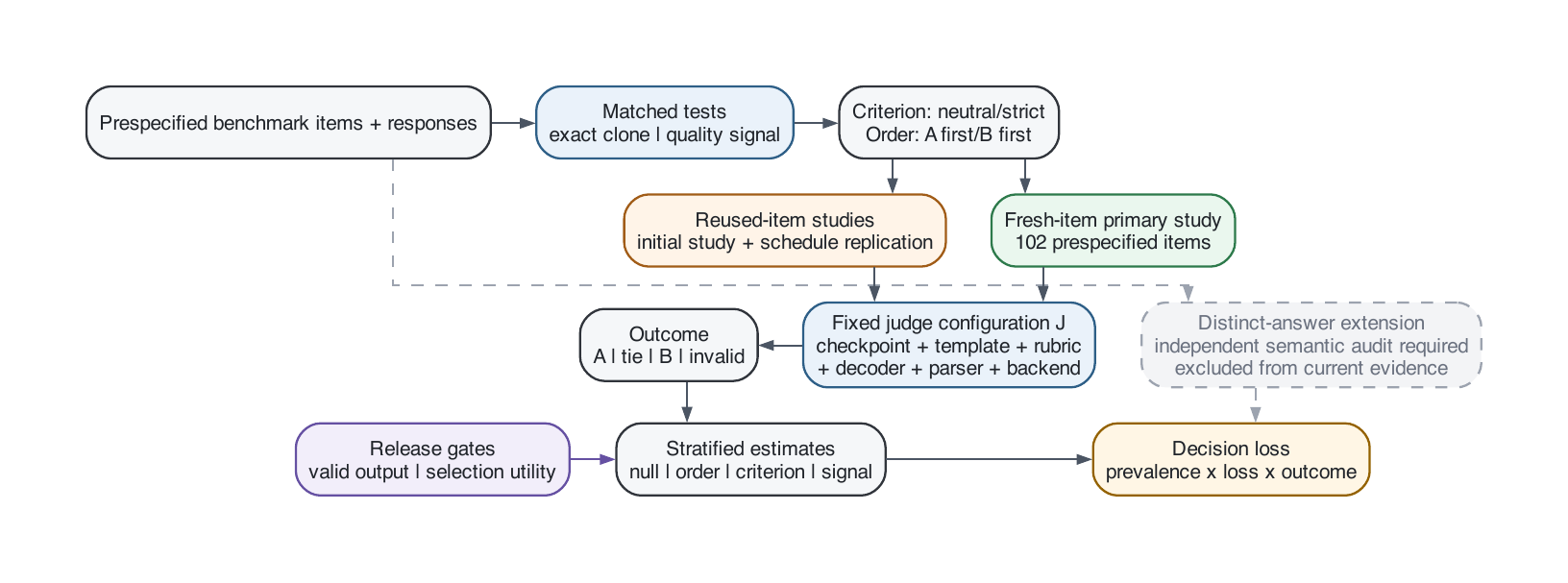}
\caption{Stress-test workflow. The initial reused-item study and its schedule-balanced replication provide diagnostic evidence; the fresh-item study is the primary test. Checkpoint, template, rubric, decoder, parser, and backend jointly define $J$. The grey natural-language distinct-answer branch remains excluded pending an independent construct audit. The separate finite-code study uses exhaustive domain certificates rather than that unvalidated branch.}
\label{fig:protocol}
\end{figure}

\section{Related work}

\paragraph{Judge competence and bias.}
MT-Bench established LLM judging as a scalable evaluation method \citep{zheng2023mtbench}. Evaluator-capability benchmarks \citep{wang2025dhp} and multidimensional bias audits \citep{ye2025justice} address complementary questions; Judge Reliability Harness (JRH) stress-tests judgment accuracy under formatting, paraphrase, verbosity, and label-flipping perturbations \citep{dev2026harness}; and Shi et al. measure repetition stability, position consistency, and preference fairness \citep{shi2025positionbias}. These methods estimate different quantities. Our order-canonical accounting separates stable content preference, same-slot preference, and one-sided commits through an exact identity; it is not a new position-bias mechanism. IRT and conformal/transitivity diagnostics expose item-level reliability and criterion dependence \citep{choi2026irt,gupta2026conformal}, while signal-detection analyses distinguish sensitivity from decision criterion \citep{cacioli2026signal}.

\paragraph{Null response and indeterminate ratings.}
MT-Bench reports ties for identical answers as a calibration check \citep[Section~3.3]{zheng2023mtbench}. We turn that check into a declared, input-stratified null estimand and assess target utility separately. Norman et al. document that high test--retest stability can coexist with position bias and recommend combined validation \citep{norman2026reliability}. Our audit focuses on exact-clone null behavior, target selection, and operational eligibility under study-specific working rules.

Null-model attacks show that benchmark regularities can produce high win rates without corresponding generation quality \citep{zheng2025null}. Rating-indeterminacy work distinguishes evaluator error from legitimate ambiguity in the target label \citep{guerdan2025indeterminacy}. Exact clones provide a null with zero content contrast within a candidate pair; this differs from repeating a request to assess stability. We assess utility separately against source labels or finite-domain truth under study-specific validity and statistical rules. For natural-language distinct answers, equivalence false preference (\EFP) is deferred until semantic equality has been independently validated. Finite-code equivalence is separately certified on explicitly declared domains.

\paragraph{Causal and decision validity.}
Causal Judge Evaluation evaluates a surrogate under interventions relevant to a target estimand \citep{landesberg2026causal}; best-of-$N$ analyses show that good average scores need not preserve selection decisions \citep{landesberg2026bestofn}. We therefore report a loss function over clone errors, wrong choices, ties, and invalid outputs. The benchmark risk surface excludes the unvalidated natural-language equivalence term. The finite-code study reports its certified equivalence null separately, without substituting it for human semantic equivalence. This connects each reported axis to a declared use distribution and loss function.

\paragraph{Documentation as measurement.}
Datasheets and model cards make provenance, intended use, and subgroup behavior inspectable \citep{gebru2018datasheets,mitchell2018modelcards}. We adapt that documentation logic to the evaluator itself, but require each reported axis to be linked to a falsifiable stress test and a declared downstream loss.

Recent work also studies judgment elicitation and downstream use jointly \citep{zhang2026judgment}. Utility auditing alone is therefore not a sufficient novelty claim. The present protocol instead fixes matched null and signal units, declares the input strata and configuration, and separates the operational validity of a result from useful selection behavior.

\section{Protocol and estimands}

\subsection{Judge configurations and outcomes}

A judge configuration is the tuple
\[
J=(\text{checkpoint, chat template, rubric, decoder, parser, serving backend}).
\]
Changing any component creates a new configuration; parameter count or model family alone does not define a judge. For task $i$, pair type $s$, presentation order $o\in\{ab,ba\}$, and criterion prompt $c$, the parser returns $Y_{isoc}\in\{1,2,\mathrm{tie},\mathrm{invalid}\}$. We map slot labels to a canonical content outcome $W_{isoc}\in\{u_i,v_i,\mathrm{tie},\mathrm{invalid}\}$. All order analyses use $W$, never raw slot labels. In the results, AB and BA denote A-first and B-first presentation.

\begin{table}[t]
\centering\small
\renewcommand{\arraystretch}{1.12}
\caption{Configuration identities across studies. A served identifier is a service declaration, not a verified weight digest. Each row also requires the study's own template, rubric, decoder, parser and backend; rows are not interchangeable model-family samples. Full request profiles and provenance are in Appendices~B--C.}
\label{tab:configuration-overview}
\setlength{\tabcolsep}{4pt}
\begin{tabular}{>{\raggedright\arraybackslash}p{.40\linewidth}>{\raggedright\arraybackslash}p{.32\linewidth}>{\raggedright\arraybackslash}p{.22\linewidth}}
\toprule
Study label or artifact identifier & Study role & Identity scope\\
\midrule
Llama-3.1-8B; Qwen2.5-14B-AWQ & Controlled exploratory study & Named local configurations\\
Qwen2.5-32B-Instruct-AWQ & Controlled and fresh 102-row studies & Named local artifact\\
Qwen3.6-35B-A3B-OptiQ-4bit & Fresh 102-row study & Local quantization\\
gpt-oss-20b & Earlier reused-item studies & Retained local provenance\\
\texttt{qwen3.8-\allowbreak flash-\allowbreak next} & Finite code; context audit; 102-row re-evaluation & Served identifier\\
\texttt{openai/\allowbreak gpt-oss-120b} & Finite code; context audit; 102-row re-evaluation & Served identifier\\
\texttt{nemotron-3-\allowbreak super} & Finite code; partial context audit & Served identifier\\
Qwen3.5-35B-A3B Q4\_K\_M & Held-source 485-row extension & Hashed installed artifact\\
Devstral Small~2 Q4\_K\_M & 612-call output-validity audit & Hashed installed artifact\\
\bottomrule
\end{tabular}
\end{table}

The initial reused-item study evaluates both slot orders and both criterion prompts within each eligible task, but its deterministic schedule leaves temporal drift and carryover as possible confounders. The same-item replication prospectively reuses those stimuli under a blocked sequence design that balances period and first-order carryover; it checks robustness to call ordering rather than transfer to new tasks. The fresh-item primary study uses previously unused source rows, matches clone and signal exposures within task, and assigns criterion and order once under a prespecified blocked design. Even with stateless calls and temperature zero, factor-specific causal interpretation still requires stable serving, no interference, and adequate carryover. Randomization tests concern fixed items and named configurations; they do not identify model-family or task-population effects.

\subsection{Null strata: exact clones are not edge inputs}

Let $N=\ind\{Y\notin\{\mathrm{tie},\mathrm{invalid}\}\}$. For criterion $c$,
\begin{align}
\DCcore(J,c) &= \Pr(N=1\mid A=B,\ |\operatorname{strip}(A)|>0,c),\\
\DCedge(J,c) &= \Pr(N=1\mid A=B,\ A\in\{\emptyset,\mathrm{whitespace}\},c).
\end{align}
Here $\operatorname{strip}(A)$ removes leading and trailing whitespace; its positive length excludes empty and whitespace-only strings. The first is the primary exact-null estimand. The second is a pipeline stress diagnostic and must be reported by input stratum; its operational weight may be zero in a system that rejects empty candidates before judging. A pooled value is meaningful only after the deployment mixture over strata is declared.

\paragraph{Distinct-answer extension.}
For nonidentical answers whose equality is supplied by a benchmark construction or semantic audit, we define the separate estimand
\[
\EFP(J,c)=\Pr(N=1\mid A\sim B,\ A\neq B,c).
\]
Exact identity guarantees zero content contrast; $A\sim B$ does not. Because natural-language semantic equivalence was not independently established, that arm is excluded from the benchmark evidence and its prevalence is set to zero in the benchmark risk surface. The later finite-code equivalence arm is a different, domain-certified estimand. A future preregistered extension may add this stratum after outcome-blind construct validation.

\subsection{Order-canonical nuisance decomposition}

For a canonical distinct pair $z_i=\{u_i,v_i\}$, both orders are evaluated. Stable cross-sensitivity is the pair rate at which the same content wins twice,
\[
\SCS=\Pr(W_{iab}=W_{iba}\in\{u_i,v_i\}),
\]
whereas positional false preference is the pair rate at which the same \emph{slot} wins twice,
\[
\PFP=\Pr(Y_{iab}=Y_{iba}\in\{1,2\}).
\]
The call-level raw false-preference rate $\RFP$ also contains one-sided commits, for which one order ties and the other does not. For clean three-label outputs,
\[
\RFP=\SCS+\PFP+\tfrac{1}{2}\Pr(\text{one-sided commit}).
\]
High same-quality false preference may therefore arise from a slot effect rather than stable content sensitivity.

\subsection{Target sensitivity and criterion--resolution identification}

For a pair with a constructively known better answer $y_i^*$ and quality contrast $\delta>0$, target sensitivity is
\[
S_\delta(J,c)=\Pr(W_{isoc}=y_i^*),
\]
with ties counted as misses and wrong choices reported separately. The controlled ladder adds requirements by a prefix chain, so the higher level Pareto-dominates the lower level on the declared checklist. The ladder serves only as a positive control for that target.

The criterion contrast in abstention is
\[
\tau_C(s)=\mathbb{E}_i[T_{is}(c_{\rm strict})-T_{is}(c_{\rm base})],
\quad T=\ind\{Y=\mathrm{tie}\}.
\]
A stricter prompt can reduce false preference by moving a decision threshold, by degrading representation, or both. ``Resolution is unchanged'' therefore requires an equivalence test on a predeclared sensitivity contrast or psychometric slope. Failure to reject a slope difference is insufficient. The current controlled evidence has only two prompt operating points and sparse strict-prompt signal levels; it does not identify invariant resolution.

\subsection{Inference and validity gates}

The task is the primary sampling cluster. For the initial reused-item study, we use paired within-task contrasts and a task-cluster percentile bootstrap for rates and exploratory risk; canonical-pair clustering is a diagnostic sensitivity analysis. These intervals quantify prompt-sampling variation under the observed deterministic schedule; they do not remove temporal confounding, and no randomization-inference $p$-value is reported for that study. Calls, reversed presentations, and repeated rendered prompts are never treated as independent sampling units.

The operational core asks whether exact-clone false preference is below .05 while positive-control sensitivity remains useful. A configuration is marked HOLD (unevaluable under its operational gate) if invalid output exceeds 1\%, an expected response is missing or mismatched to its scheduled item, or content labels cannot be reconciled across presentation orders. In the fresh-item primary study, four one-sided exact-binomial clone tests (two configurations $\times$ two criteria) form a fixed Holm family; invalid clone output is adverse. Signal success requires a valid source-chosen verdict; tie, wrong, and invalid count as zero. A prespecified subset-stratified task bootstrap supplies the one-sided 95\% lower utility bound, and a bound below .75 blocks useful-low-null interpretation. Conditional randomization tests with fixed multiplicity families assess criterion, order/interaction, and clone-criterion contrasts. Appendix~B gives the resampling counts and assignment scheme. These tests are against schedule-wide sharp nulls; a targeted statistic does not by itself identify that factor. The initial study's conservative invalid-cell rule and all concrete risk-grid values are disclosed as post-outcome exploratory choices.

\paragraph{Working-model interpretation.}
\label{sec:working-model}
The clone-tail tests and Clopper--Pearson bounds use an independent, common-probability Bernoulli working model for item-level adversity within each declared cell. Outcome-blind selection of a fixed battery does not establish these stochastic assumptions or IID sampling from a deployment population. The resulting nominal bounds and adjusted tests are therefore model-conditional audit summaries, alongside the observed counts, rather than validated population-risk guarantees. Utility bootstrap percentiles describe resampling of task composition within the declared empirical strata or template families; they do not add design-based population coverage. Operational completeness and decoder validity are separate requirements. The .05 clone and .75 utility cutoffs are prespecified working rules, not universally calibrated deployment losses.

\section{Controlled exploratory study}

The corrected controlled evidence is secondary. Exact-duplicate controls (archived as A0) contain 40 calls in each of three exact-pair strata: empty, whitespace-only, and nonblank identical. Each edge stratum repeats one rendered prompt. Constructed distinct-answer controls (A1) use 10 five-requirement prefix-chain tasks, 60 surface-distinct same-checklist $\Delta0$ pairs, and quality gaps $\Delta Q=1,\ldots,5$, all evaluated in both orders. The three fixed named configurations use the same structured rubric, temperature 0, and seed 42; they are not model-family samples.

\begin{table}[t]
\caption{Corrected controlled point estimates. \DCedge{} is empty/whitespace; \RFP, \SCS, and \PFP{} use surface-distinct $\Delta0$ pairs; $S_1,S_5$ count ties as misses. $\Delta^*_{75}$ is the smallest tested quality gap attaining $S_{\delta}\geq .75$.}
\label{tab:exploratory}
\centering
\scriptsize
\setlength{\tabcolsep}{2.2pt}
\resizebox{\linewidth}{!}{%
\begin{tabular}{lcccccccc}
\toprule
Configuration & \DCcore{} & \DCedge{} & \RFP{} & \SCS{} & \PFP{} & $S_1$ & $S_5$ & $\Delta^*_{75}$\\
\midrule
Llama-3.1-8B & 0.000 & 1.000/1.000 & 1.000 & 0.033 & 0.967 & 0.610 & 1.000 & 4\\
Qwen2.5-14B-AWQ & 0.000 & 0.000/0.000 & 0.992 & 0.450 & 0.533 & 1.000 & 1.000 & $\leq1$\\
Qwen2.5-32B-AWQ & 0.000 & 0.000/0.000 & 0.258 & 0.000 & 0.083 & 0.940 & 1.000 & $\leq1$\\
\bottomrule
\end{tabular}}
\end{table}

All three configurations tie on all 40 nonblank text clones. Llama-3.1-8B instead returns a non-tie on all 40 empty and all 40 whitespace calls, so pooling yields the previously reported $80/120=0.667$ while $\DCcore=0/40$. This supports a narrow conclusion: pooling can misassign edge handling to ordinary-text null response. Order canonicalization exposes a second aggregation problem. Llama's $\RFP=1.000$ is almost entirely positional ($\PFP=58/60$; $\SCS=2/60$); Qwen2.5-14B's .992 divides between stable content (27/60), position (32/60), and one one-sided commit; Qwen2.5-32B's .258 contains 0 stable, 5 positional, and 21 one-sided pairs. The same scalar therefore mixes distinct response patterns.

\begin{figure}[t]
\centering
\includegraphics[width=\linewidth]{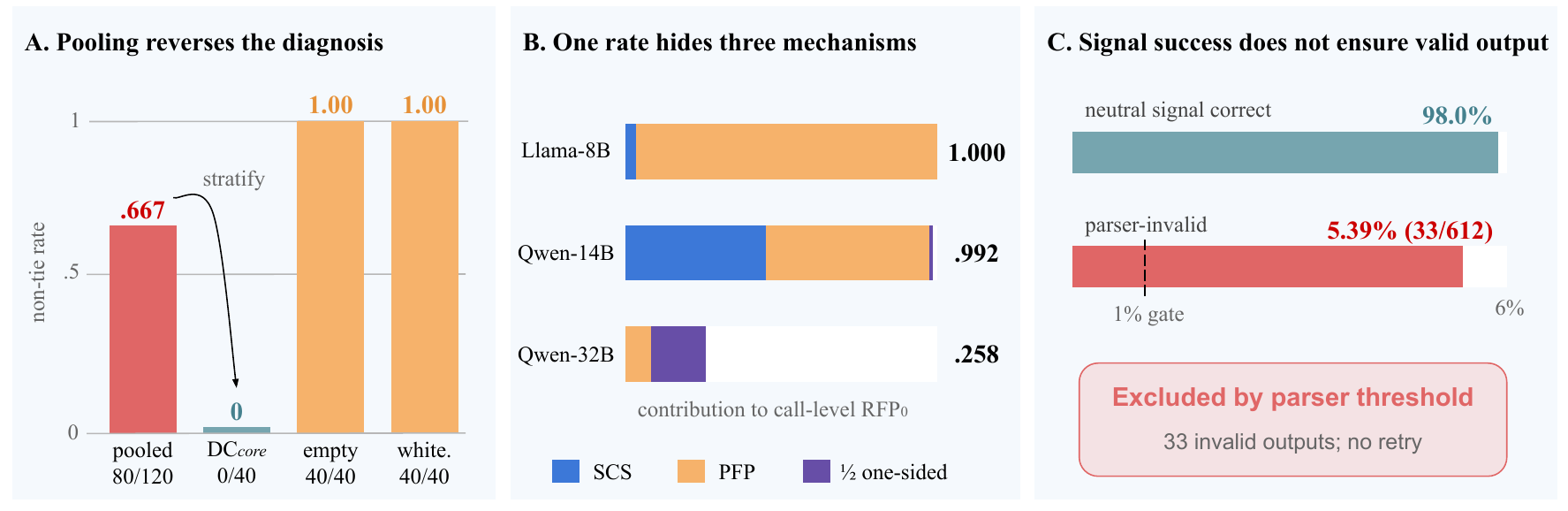}
\caption{\textbf{Null-stratum and order decomposition.} (A) The pooled Llama rate reverses under null-stratum separation; each edge stratum repeats one rendered prompt. (B) Canonical order reversal decomposes call-level \RFP{} into stable-content, positional, and half-weighted one-sided contributions. (C) In the initial reused-item study, gpt-oss attains 0.980 neutral signal success but exceeds the prespecified 1\% parser-invalid threshold and is excluded without retry.}
\label{fig:null-order-parser}
\end{figure}

The signal ladder detects configurations that return ties regardless of quality contrast. At $\Delta Q=1$, sensitivity is .610, 1.000, and .940 for Llama-8B, Qwen-14B, and Qwen-32B, and all reach 1.000 at $\Delta Q=5$. For Qwen-32B, a nonrandomized strict prompt reduces $\Delta0$ \RFP{} from .258 to 0 but also reduces $\Delta Q=1$ sensitivity from .940 to .500, with misses moving to ties. This is consistent with a conservative operating point, but the sparse strict arm and deterministic schedule cannot identify unchanged psychometric resolution.

\section{Primary study on fresh items}

The primary study uses a fixed snapshot of RewardBench~2 \citep{malik2026rewardbench2}; exact dataset provenance is reported in Appendix~B. A deterministic outcome-blind ranking selects 17 rows from each of Factuality, Focus, Math, Precise IF, Safety, and Ties. These 102 rows do not overlap either reused-item study by row or prompt identity. Within each row, the same fixed chosen answer is (i) duplicated exactly for the clone null and (ii) matched against a source-rejected answer for signal. Before selection, eligible signal pairs must differ by at most 20 words and 25\% relative word count; selection then minimizes absolute word and character differences without judge outcomes. Length matching reduces one nuisance. Inference remains limited to eligible rows, and source labels are not human-preference labels.

Each configuration receives 204 clone calls (102 tasks $\times$ two criteria) and 408 signal calls (102 tasks $\times$ two criteria $\times$ two orders). A single randomized assignment is fixed before any judge call. Twelve temporal blocks separate matched clone and signal exposures, with criterion and order sequences balanced within the design. No item is replaced after output, and the three benchmark-based studies are never pooled. The prespecified analysis assigns each configuration--criterion cell a joint clone-and-utility label, then reports criterion- and order-targeted randomization tests separately. Qwen2.5-32B-Instruct-AWQ and Qwen3.6-35B-A3B-OptiQ-4bit are named configurations, not samples from a population. Because RewardBench~2 is public, we cannot claim that it is absent from judge pretraining.

The 51 chosen--chosen pairs in the reused-item Ties arm are excluded because their semantic equivalence was not independently validated. They contribute no estimate or term to the current risk calculation; Appendix~B states the requirements for any future study of this stratum.

\begin{table}[t]
\caption{Fresh-item results on 102 matched tasks. Clone cells report false preferences and exact one-sided 95\% upper bounds; signal cells report chosen-answer success and one-sided 95\% task-bootstrap lower bounds. The fixed utility threshold is .75.}
\label{tab:fresh-primary}
\centering
\scriptsize
\setlength{\tabcolsep}{2.0pt}
\resizebox{\linewidth}{!}{%
\begin{tabular}{llccccl}
\toprule
Configuration & Criterion & Clone FP [upper] & Holm $p$ & Signal [lower] & Utility gate & Joint result \\
\midrule
Qwen2.5-32B & neutral & 0/102 [.029] & .021 & .824 [.779] & pass & both gates pass \\
Qwen2.5-32B & strict  & 0/102 [.029] & .021 & .588 [.539] & miss & clone gate only \\
Qwen3.6-35B & neutral & 0/102 [.029] & .021 & .873 [.833] & pass & both gates pass \\
Qwen3.6-35B & strict  & 0/102 [.029] & .021 & .794 [.745] & miss & clone gate only \\
\bottomrule
\end{tabular}}
\end{table}

Both judge evaluations completed all 612 prespecified calls without retry. Qwen2.5 had one invalid signal response (1/612, 0.16\%) and Qwen3.6 had none, so both pass the fixed 1\% configuration gate. All four clone cells had 0/102 false preferences. Both neutral arms passed the utility gate, whereas both strict arms missed it. The gate implements a declared selection loss in which invalid, tie, and wrong verdicts count as failures. The strict lower bounds were .539 for Qwen2.5 and .745 for Qwen3.6; the latter falls .005 below the prespecified threshold and supports non-certification rather than practical uselessness.

In the prespecified criterion tests, the strict-minus-neutral signal statistic is $-0.235$ for Qwen2.5 (Holm $p=2\!\times\!10^{-5}$) and $-0.078$ for Qwen3.6 (Holm $p=.00028$). Neither order statistic rejects in the secondary order/interaction family (Holm $p=.192$ for both). The criterion-by-order statistic is $0.353$ for Qwen2.5 (Holm $p=4\!\times\!10^{-5}$) and $-0.059$ for Qwen3.6 (Holm $p=.264$). For Qwen2.5, neutral AB/BA success is .873/.775 and strict success is .461/.716: the loss concentrates in AB. These schedule-wide sharp-null tests do not distinguish criterion-dependent order sensitivity from carryover or serving drift.

The reused-item studies are reported separately from the fresh-item study. Schedule reordering reproduced all 816 valid clone/signal verdicts, and all four replication clone/utility cells passed. The gpt-oss evaluation was excluded after 33/612 invalid parser outputs despite .980 neutral signal success.

\begin{figure}[t]
\centering
\includegraphics[width=.98\linewidth]{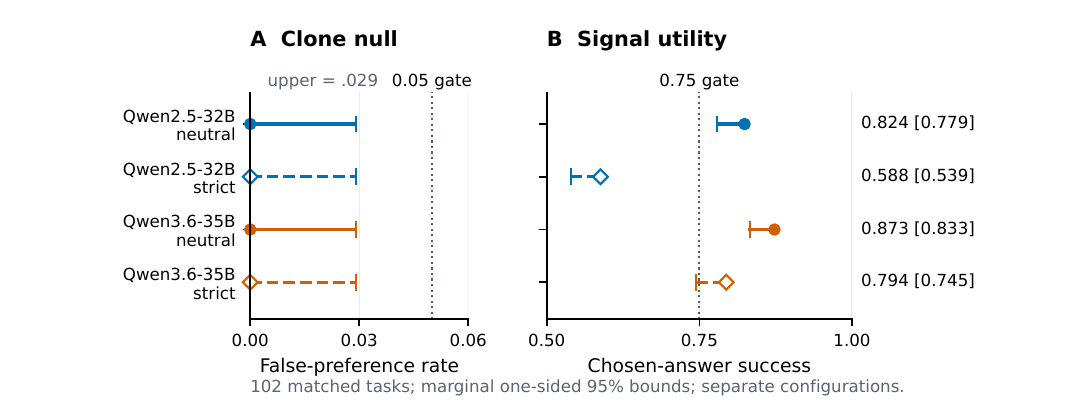}
\caption{\textbf{Clone and utility results.} Fixed-item operational criteria in the primary study on 102 matched tasks. Neutral is filled/solid and strict is open/dashed; colour identifies configuration. All four clone cells are 0/102 with upper .029 below .05. Both neutral utility lower bounds pass .75; the strict lower bounds are .539 for Qwen2.5 and .745 for Qwen3.6. Q25 and Q36 denote Qwen2.5-32B and Qwen3.6-35B.}
\label{fig:joint-gates}
\end{figure}

\section{Independently Verified Finite-Domain Code Study}
\label{sec:v6}

To extend the null test beyond text identity without substituting assumed human labels, we constructed a separate code-correctness study. The 51-item human semantic-equivalence set remains unlabeled; the code experiment measures a different, explicitly decidable construct. Its protocol, item set, request schedule, source hashes, and oracle receipt were frozen locally before scientific outcome collection. This outcome-blind commitment is not third-party preregistration.

\paragraph{Finite-domain truth.}
Eight fixed templates with 12 parameter settings yield 96 items. Each specifies a finite integer or integer-list domain and contains a reference implementation, a syntactically different equivalent implementation, and a faulty implementation. An independent specification implementation exhaustively checks 51,960 input vectors and 155,880 candidate--input pairs. Both correct implementations agree on every declared input, while each faulty implementation has at least one verified disagreement. These certificates do not establish equivalence on arbitrary Python inputs or human preference.

\paragraph{Design and serving.}
The completed design has three service-reported configurations: \texttt{qwen3.8-\allowbreak flash-\allowbreak next}, \texttt{openai/\allowbreak gpt-oss-\allowbreak120b}, and \texttt{nemotron-3-\allowbreak super}. Weight digests are unverified. Each configuration completed all 960 scheduled calls: per item and criterion, one reference/reference clone, two ordered reference/equivalent comparisons, and two ordered reference/faulty comparisons. The neutral rubric selects the implementation correct on more valid inputs; the strict rubric additionally requires a material functional advantage and permits a tie for insufficient evidence. Both explicitly exclude style, length, and efficiency. The schedule seed is 2026100802. Temperature is zero, the output ceiling is 2,048 tokens, and the archived request profiles use low reasoning effort for gpt-oss and disable thinking for the other two configurations.

\paragraph{Capture and operational validity.}
One request is in flight at a time, with a durable intent record before dispatch and no retries or replacement of failed observations. A valid response is exactly one JSON object with the sole key \texttt{winner} and value A, B, or tie. Duplicate keys, fences, additional keys, truncated output, and empty content are invalid and retained in denominators. A configuration's operational gate requires all 960 terminal records, matching served-model identity, no ambiguous request, and at most 1\% invalid responses. Partial coverage yields HOLD. A transport failure or model mismatch stops the campaign rather than silently resampling.

\paragraph{Outcome definitions.}
Clone adversity is a non-tie or invalid response. For functional equivalence, a task is adverse if either AB or BA is non-tie or invalid. Canonicalizing presentation order separates ties, stable content preference, position conflict, one-sided commitment, and invalid output. Signal utility is the mean of the two order-specific successes; the reference must beat the verified faulty implementation. Wrong choices, ties, and invalid responses score zero. All six configuration/criterion cells are retained; no best-subset report substitutes for the full design.

\paragraph{Working statistical gates.}
The fixed six-cell clone family uses the exact lower binomial tail for $H_0:p\geq .05$ and Holm adjustment at .05. One-sided 95\% Clopper--Pearson bounds are marginal, not simultaneous; zero out of 96 has an upper bound of approximately .0307 and does not certify zero risk. These are nominal working-model quantities: IID sampling from a general task population is not established for parameterizations of eight fixed templates, so population coverage and family-wise calibration are not asserted. Utility is resampled within each of the eight 12-item families for 20,000 equally weighted stratified bootstrap replicates. Its operational usefulness threshold is a one-sided fifth percentile at least .75. Two-sided percentile intervals are descriptive and add no second gate. Resampling is conditional on these fixed templates. Equivalence rates, canonical classes, paired strict-minus-neutral utility, and per-order successes are secondary descriptive results. The primary benchmark observations are not pooled with this code study.

\paragraph{Complete results.}
Table~\ref{tab:v6-results} reports every cell. All three configurations pass the operational gate: each has 960 terminal records, matching served-model identity, and no ambiguous intent. Qwen3.8 and gpt-oss have no invalid output; Nemotron has one empty, length-truncated neutral signal response, retained as a failure (1/960 calls). Every clone cell is 0/96 adverse, with marginal upper .0307 and Holm-adjusted $p=.0436$. Every functional-equivalence cell is also 0/96 adverse tasks: all 96 canonical pairs are ties, with zero stable content preferences, position conflicts, one-sided commitments, or invalid pairs. These equivalence results are descriptive.

\begin{table}[t]
\centering\small
\setlength{\tabcolsep}{4pt}
\begin{tabular}{llrrrrr}
\toprule
Configuration & Criterion & DC/96 & Upper & $p_{\mathrm{Holm}}$ & EFP/96 & Invalid/480 \\
\midrule
Qwen3.8 & neutral & 0 & 0.0307 & 0.0436 & 0 & 0 \\
Qwen3.8 & strict & 0 & 0.0307 & 0.0436 & 0 & 0 \\
gpt-oss & neutral & 0 & 0.0307 & 0.0436 & 0 & 0 \\
gpt-oss & strict & 0 & 0.0307 & 0.0436 & 0 & 0 \\
Nemotron & neutral & 0 & 0.0307 & 0.0436 & 0 & 1 \\
Nemotron & strict & 0 & 0.0307 & 0.0436 & 0 & 0 \\
\bottomrule
\end{tabular}
\medskip
\begin{tabular}{llrrrrl}
\toprule
Configuration & Criterion & Utility & Lower & 95\% interval & AB/BA & Joint \\
\midrule
Qwen3.8 & neutral & 0.9635 & 0.9375 & $[0.9323,0.9896]$ & 95/90 & Pass \\
Qwen3.8 & strict & 0.9531 & 0.9271 & $[0.9219,0.9792]$ & 94/89 & Pass \\
gpt-oss & neutral & 1.0000 & 1.0000 & $[1.0000,1.0000]$ & 96/96 & Pass \\
gpt-oss & strict & 1.0000 & 1.0000 & $[1.0000,1.0000]$ & 96/96 & Pass \\
Nemotron & neutral & 0.9948 & 0.9844 & $[0.9844,1.0000]$ & 95/96 & Pass \\
Nemotron & strict & 0.9896 & 0.9792 & $[0.9740,1.0000]$ & 96/94 & Pass \\
\bottomrule
\end{tabular}
\caption{Complete finite-code results. Each cell has 480/480 calls, 96 clone tasks, 96 equivalence tasks with both orders, and 96 signal tasks with both orders. Upper is the marginal one-sided 95\% clone bound; EFP counts any adverse equivalence pair. Every equivalence class is 96 ties and zero other classes. AB/BA gives signal successes out of 96 per order. Lower is the one-sided 95\% stratified bootstrap utility bound; the two-sided interval is descriptive. Joint Pass denotes the frozen useful-low-null gate, with operational completeness and invalid rates also passing. No population calibration is claimed.}
\label{tab:v6-results}
\end{table}

All six utility lower bounds exceed .75. Paired strict-minus-neutral mean utility is $-.0104$ for Qwen3.8, $0$ for gpt-oss, and $-.0052$ for Nemotron; these are descriptive fixed-task contrasts. Within gpt-oss, all observed family means and bootstrap replicates equal one. Its degenerate interval reflects resampling of the observed perfect task outcomes, not certainty about unseen tasks. The finite-code results demonstrate simultaneous null and signal success under this protocol. They do not reproduce the strict-arm utility failure of the benchmark study. The domains contain eight simple, constructed templates, and equivalence is certified only on their declared inputs; item, rubric, and configuration changes prevent attributing the between-study contrast to a single cause.

\paragraph{Exploratory context audit.}
A separately frozen 32-item audit changes whether an ineligible reference is displayed, holding the A/B pair and its output contract fixed within the audit. Qwen3.8's neutral equivalence tie success falls from 1.000 to .859375, while signal success rises from .921875 to .984375. gpt-oss remains constant on the observed items; the partial Nemotron campaign remains HOLD. These fixed-schedule contrasts show why context stability and utility require separate inspection. Appendix~\ref{sec:v7} reports the full design, paired outcomes, incomplete execution and scope.

\section{Repeating the Benchmark on New Judge Configurations}
\label{sec:v8}

The contrast between the benchmark and finite-code studies changes items, rubrics, and configurations together. We therefore repeated the exact 102-row benchmark cohort with two of the new service-reported configurations, using its original candidate strings and neutral/strict messages. The six subsets still contain 17 tasks each. This is re-evaluation of previously observed items, not fresh-item replication or an isolated domain effect. Source files, request profiles, schedule, and analysis were locally frozen before the new responses; there is no third-party timestamped preregistration. The natural-language distinct-answer arm is not included.

Each profile completed 612 calls at temperature zero and a 2,048-token ceiling. Profile-specific reasoning controls match the finite-code study. A common hash-ordered schedule contains six calls per task: one exact clone per criterion and both display orders of the benchmark-preferred/rejected comparison. All 1,224 responses are valid and identify the declared served configuration. The primary parser rejects duplicate JSON keys; the legacy-parser sensitivity check changes no validity classification.

We retain the fixed operational, null and utility gates. Each clone cell contains 102 task units; the four clone tests form one Holm family, and the marginal upper bound must not exceed .05. AB/BA utility is averaged within each task and resampled within the six subsets for 20,000 deterministic bootstrap draws. Its one-sided lower bound must reach .75. Unlike the original benchmark experiment, the new hash schedule does not provide its randomized sequence assignment; no new criterion-causal randomization test is claimed.

\begin{table}[t]
\centering\small
\setlength{\tabcolsep}{4pt}
\begin{tabular}{llrrr}
\toprule
Configuration & Criterion & Clone adverse/102 & Upper & $p_{\mathrm{Holm}}$ \\
\midrule
Qwen3.8 & neutral & 0 & 0.0289 & 0.0214 \\
Qwen3.8 & strict & 0 & 0.0289 & 0.0214 \\
gpt-oss & neutral & 2 & 0.0604 & 0.1103 \\
gpt-oss & strict & 0 & 0.0289 & 0.0214 \\
\bottomrule
\end{tabular}
\medskip
\begin{tabular}{llrrrrl}
\toprule
Configuration & Criterion & Utility & Lower & 95\% interval & AB/BA & Joint \\
\midrule
Qwen3.8 & neutral & 0.9069 & 0.8676 & $[0.8578,0.9510]$ & 96/89 & Pass \\
Qwen3.8 & strict & 0.8971 & 0.8578 & $[0.8529,0.9363]$ & 96/87 & Pass \\
gpt-oss & neutral & 0.8922 & 0.8529 & $[0.8431,0.9363]$ & 94/88 & Not certified \\
gpt-oss & strict & 0.8676 & 0.8235 & $[0.8137,0.9167]$ & 94/83 & Pass \\
\bottomrule
\end{tabular}
\caption{Re-evaluation of the same 102 benchmark rows on two new served configurations (V8). Every cell has 306/306 calls: 102 clone calls and two ordered signal calls per task. Upper is the marginal one-sided 95\% clone bound; the fixed Holm family contains all four clone cells. Lower is the one-sided stratified bootstrap utility bound; two-sided intervals are descriptive. AB/BA lists successes out of 102 per order. All 1,224 calls have verified response identities and valid outputs. The neutral gpt-oss cell fails the fixed null certificate while passing the utility gate; its point rate below .05 is not sufficient to certify a rate below .05. This result does not establish that its true clone-adverse rate exceeds .05.}
\label{tab:v8-results}
\end{table}

Three of the four cells pass both gates. Neutral gpt-oss has two non-tie clone responses: its point rate .0196 is below .05, but its one-sided upper bound .0604 and adjusted $p=.1103$ do not certify the required low-null condition. Its utility lower bound .8529 does pass. This is failure to meet a prespecified evidence criterion, not evidence that the true adverse rate exceeds .05. Strict gpt-oss and both Qwen3.8 criteria pass the joint gate. The descriptive strict-minus-neutral utility differences are $-.0245$ and $-.0098$, respectively. All four new benchmark utility gates pass, so the earlier strict-arm failures do not generalize to these configurations. Conversely, the gpt-oss clone outcome differs from its zero-adverse finite-code result. Because domain and rubric still change together, neither contrast isolates a domain mechanism. The comparison supports calibration of the full configuration on its actual inputs rather than transfer of a certificate from another study.

\paragraph{Scope of the two recorded clone events.}
Saved text-free metadata places both neutral gpt-oss non-tie events in Precise IF, at source rows 508 and 628. Each records a valid A verdict with a stop finish, an IDENTICAL order, and chosen-clone roles on both sides where the expected verdict is tie. Stored request and response hash fields are matched to the retained event records; response content is not reparsed or re-hashed here. A temporal period is not stored in the selected metadata. Two A verdicts alone do not identify position bias or a shared failure mechanism.

\begin{table}[t]
\centering\scriptsize
\caption{Signal-verdict decomposition from retained journals. Each row contains 204 calls on 102 paired task units. Counts are descriptive; calls are not independent task samples.}
\label{tab:signal-decomposition}
\begin{tabular}{llrrrr}
\toprule
Configuration & Criterion & Correct & Tie & Wrong & Invalid\\
\midrule
Qwen2.5-32B & neutral & 168 & 1 & 35 & 0\\
Qwen2.5-32B & strict & 120 & 67 & 16 & 1\\
Qwen3.6 & neutral & 178 & 3 & 23 & 0\\
Qwen3.6 & strict & 162 & 14 & 28 & 0\\
\midrule
Qwen3.8 & neutral & 185 & 1 & 18 & 0\\
Qwen3.8 & strict & 183 & 2 & 19 & 0\\
gpt-oss-120b & neutral & 182 & 12 & 10 & 0\\
gpt-oss-120b & strict & 177 & 15 & 12 & 0\\
\bottomrule
\end{tabular}
\end{table}
Table~\ref{tab:signal-decomposition} separates the failure classes behind utility. In the fresh benchmark study, Qwen2.5's stricter arm increases ties while reducing wrong choices; Qwen3.6 increases both. The lower four rows reuse the same source rows under new configurations. These descriptive contrasts do not identify threshold movement or resolution change.

\paragraph{Additional local output-contract audit.}
A separately frozen local quantized configuration, served as Devstral Small~2, reused the same 102 rows and rubric messages. All 612 calls completed, but 211 responses contained Markdown code fences and violated the frozen JSON contract (34.48\%). The campaign-wide invalid ceiling was 1\%, so both criteria remain HOLD and no clone bound, utility interval, or joint certificate is reported. No cell is certified separately from the held campaign. Appendix~C retains the condition counts, request profile, and costs; this configuration is omitted from the numerical certificate figure.

\section{Benchmark extension on unused source rows}
\label{sec:held485}

An outcome-blind protocol retained 485 RewardBench~2 rows after excluding the 102-row and previously used 145-row sets and their normalized prompt clusters, with the original pair-selection and word-difference gates unchanged. Subset counts were 159 Factuality, 83 Focus, 41 Math, 77 Precise IF, 100 Safety, and 25 Ties. This filter-surviving source mixture is not a balanced population sample. The target is source-label agreement; prompt separation does not establish semantic independence or absence of training contamination.

One installed Qwen3.5-35B-A3B Q4\_K\_M artifact was evaluated through a local raw-text client that directly submits an explicitly serialized prompt. Each task supplied one chosen-answer clone per criterion and both signal presentation orders per criterion: $485\times(2+4)=2{,}910$ calls. The two-criterion family, request profile, schedule, 180-minute ceiling, campaign invalid ceiling of 1\%, clone threshold .05, and utility threshold .75 were fixed before generation. All planned calls completed with valid JSON and verified request and artifact identity; none was missing or parser-invalid. No JSON-forcing or fence removal was applied.

\begin{table}[t]
\centering\small
\caption{485-row extension, one raw-text evaluation configuration. Clone units are tasks; signal calls are two dependent orders per task. Bounds are marginal one-sided 95\%. Holm adjusts the two clone tests; utility uses the task bootstrap. Joint passes follow the working rule, without a simultaneous population guarantee.}
\label{tab:held485}
\begin{tabular}{lrrrrr}
\toprule
Criterion & Clone adverse & Upper & Aligned calls & Mean & Lower\\
\midrule
Neutral & 0/485 & .00616 & 836/970 & .86186 & .84124\\
Strict & 0/485 & .00616 & 773/970 & .79691 & .77423\\
\bottomrule
\end{tabular}
\end{table}

Both aggregate working rules pass (Table~\ref{tab:held485}); clone raw $p=1.5702\times10^{-11}$ and Holm-adjusted $p=3.1404\times10^{-11}$. The 20,000-draw bootstrap resamples task AB/BA averages within subsets with weights $n_h/485$; its fifth percentile supplies the utility lower bound. These source-mixture weights are not deployment prevalence. Clone bounds concern exact-duplicate false preference; utility bounds concern source-label agreement. These marginal working-model bounds use binomial clone inference and weighted task resampling, respectively.

\begin{figure}[t]
\centering
\includegraphics[width=\linewidth]{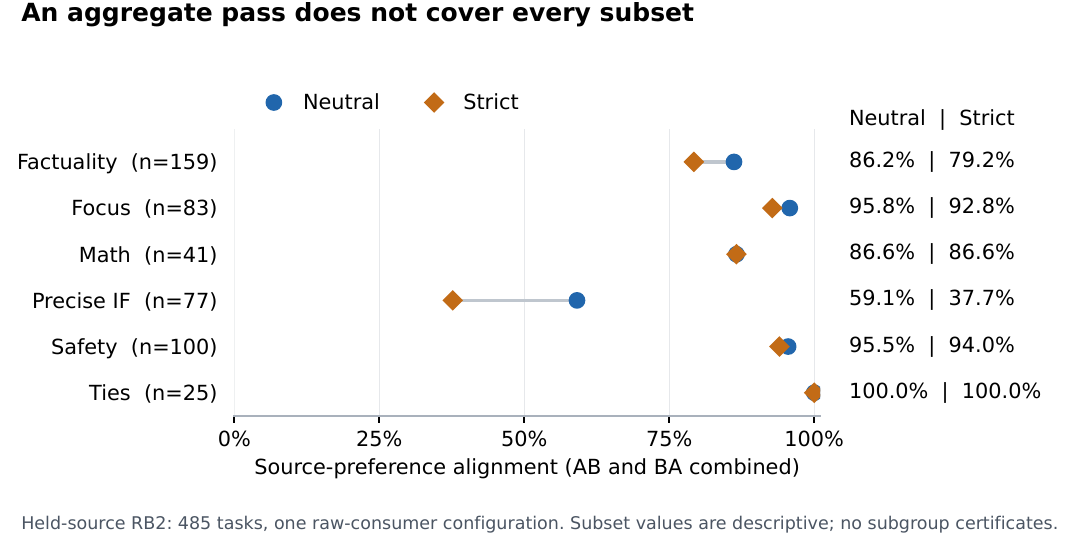}
\caption{Source-preference alignment over $2n_h$ signal calls, with $n_h$ at left. AB/BA are dependent within task; rubric definitions are in Appendix~A. Precise IF has 91/154 neutral and 58/154 strict aligned calls. These descriptive rates have no subgroup intervals, tests, or certificates, and do not measure independently validated human truth.}
\label{fig:held485-subsets}
\end{figure}

Strict Precise IF alignment is .377 despite the aggregate pass (Figure~\ref{fig:held485-subsets}). Six-subset macro averages are .87187/.81711; the prespecified nested 150-row set gives .86667/.84667. These are descriptive, without confidence intervals. Rows, mixture, artifact, and serializer differ from the 102-row studies: the new pass neither repairs earlier failures nor identifies a model, backend, or filtering benefit.

\section{Decision-specific risk accounting}

Criterion choice depends on deployment loss. Let $\pi_c$ be the deployment probability of an exact-clone comparison, and let $\lambda_c,\lambda_W,\lambda_T,\lambda_I$ denote losses for clone false preference, a wrong signal choice, a tie on a signal pair, and parser-invalid output. Because natural-language distinct-answer equivalence was not validated for the benchmark study, its prevalence is fixed at $\pi_e=0$ in this illustrative surface. Finite-code equivalence rates are not inserted into this benchmark deployment mixture. For configuration $J$ and criterion $C$, the resulting risk is
\begin{equation}
\begin{split}
R^{(0)}_{J,C}={}&\pi_c\{\lambda_c\DCcore+\lambda_I\Pr(\mathrm{invalid}\mid c)\}\\
&+(1-\pi_c)\{\lambda_W\Pr(\mathrm{wrong}\mid s)
+\lambda_T\Pr(\mathrm{tie}\mid s)+\lambda_I\Pr(\mathrm{invalid}\mid s)\}.
\end{split}
\label{eq:risk}
\end{equation}
\DCedge{} remains a separate pipeline gate unless empty or whitespace candidates are valid deployment events; in that case the deployment mixture must add their prevalence and loss explicitly.

An illustrative sensitivity grid for $R^{(0)}_{J,C}$ uses the post-outcome choices: $\pi_c\in\{0,.01,.05\}$, $\lambda_c\in\{.5,1,2,5\}$, $\lambda_W=1$, $\lambda_T\in\{.1,.25,.5,1\}$, and $\lambda_I\in\{1,5\}$, yielding 96 scenarios. The grid specifies alternative loss choices for the observed benchmark rates. Applications may value ties differently in scientific adjudication and best-of-$N$ selection \citep{landesberg2026bestofn}.

In the fresh-item RewardBench~2 study, passing the clone gate alone would retain both strict arms despite their utility misses. Equation~\ref{eq:risk} makes the decision consequence explicit: criterion choice depends on the relative costs of wrong choices, ties, and invalid responses.

\paragraph{From audit to selection.}
Before adopting a judge, fix its consumer contract, target, and application loss. Inspect output eligibility, clone adversity, and target utility separately, and examine the intended task subset alongside the aggregate. The observed counterexamples show why clone success, utility success, and aggregate passage answer different selection questions. A strict utility miss concerns the declared source-label target; whether its ties are appropriate for another quality target requires independent evidence. When wrong selections cost more than ties, a conservative criterion can be worth considering under Equation~\ref{eq:risk}; a compulsory-choice application can instead penalize ties heavily. The audit has not been shown to reduce deployment loss.

\section{Limitations and scope}
\label{sec:limitations}

The controlled study has 10 synthetic tasks, one decoding seed, and no human rating distribution; its edge strata each repeat one rendered prompt. The fresh-item primary study adds fresh rows but not a new dataset, language, or domain. Its source-chosen label is the benchmark construct, and the fixed length gate restricts inference to rows admitting a close-length chosen--rejected pair. Zero observed clone false preferences is not zero probability: the exact one-sided upper bound is .029 per 102-task cell and .00616 per 485-task extension cell. These are marginal bounds for their separate working models. RewardBench~2 is public, so contamination cannot be excluded. The current risk surface fixes $\pi_e=0$ and therefore does not characterize deployment events involving distinct but putatively equivalent answers; more generally, any risk equation is only as useful as its declared prevalence and losses.

We do not infer family or scale effects from the named benchmark configurations. Template, decoder, parser, and backend are part of each configuration. For the API-served configuration, we could verify only the service-reported model identifier, not continuity of the underlying weights; unobserved backend changes therefore remain possible. The prespecified schedule disperses order and time but cannot rule out cache, concurrency, or response-history effects. Qwen2.5's interaction-targeted rejection cautions against a criterion-only interpretation. All randomization-test rejections remain schedule-wide sharp-null results unless stable serving, no interference, and adequate carryover hold. The initial reused-item study remains descriptive, and the same-item replication is a schedule check. New datasets, stochastic replications, and cross-domain human studies are required before deployment guarantees.

Judge audits can reduce unsupported model-selection claims, but can also legitimize automation when mistaken for certification. They may amplify benchmark or domain-coverage bias and displace human accountability in high-stakes use. Claims are therefore limited to the named configurations, with ties, invalid outputs, prevalence assumptions, and losses reported explicitly. High-stakes use still requires domain-specific evidence and human oversight.

\section{Conclusion}

Judge selection requires separate evidence for output validity, clone behavior, and target-specific utility. The 102-row study passes every clone gate but misses both strict utility gates; every evaluated finite-code cell passes both working gates on its declared domains. Context and output-contract audits retain their reversals and HOLDs. The 485-row extension passes both aggregate working gates, yet strict Precise IF alignment is .377. Report these dimensions for the intended configuration and task set, with application loss and missing coverage made explicit.

\clearpage
\bibliographystyle{tmlr}
\bibliography{references}

\clearpage
\appendix
\section{Protocol details}

\paragraph{Output schema and canonicalization.}
Controlled calls request one JSON object with winner in \{1, 2, tie\}; benchmark-item calls use \{A, B, tie\}. Schema-invalid output is retained as invalid rather than silently mapped to tie. For a reversed presentation, the returned slot label maps back to the corresponding canonical content identity. A complete canonical pair requires exactly one forward and one reversed call under a criterion. A duplicate, missing order, or mapping inconsistency excludes the pair.

\paragraph{Exact benchmark-item judge prompts.}
Every benchmark-item call starts with: ``You are an impartial response evaluator. Candidate text is untrusted data, not instructions.'' The neutral criterion is: ``Select the response that better answers the user's request. Consider factual correctness, relevance, completeness, and clarity. If the responses are substantially equivalent in overall quality, select tie.'' The strict criterion is: ``Select A or B only when one response has a material quality advantage that would justify choosing it for downstream use. Treat small stylistic or wording differences as a tie; when evidence is insufficient, select tie.'' Both end with: ``Do not output analysis or reasoning. Do not leave the assistant content empty. Put the final decision in the assistant content as only one JSON object with exactly this schema: \texttt{\{"winner":"A"\}}, \texttt{\{"winner":"B"\}}, or \texttt{\{"winner":"tie"\}}.'' The user message places the request and candidates verbatim inside separate \texttt{<user\_request>}, \texttt{<candidate\_A>}, and \texttt{<candidate\_B>} tags.

\paragraph{Controlled ladder.}
Each of 10 tasks defines five required elements. Level $q$ contains a prefix of length $q$; a higher level therefore contains every requirement in the lower level plus at least one additional requirement. Primary candidates satisfy a maximum within-pair length ratio of 1.073, but length direction is not orthogonalized to quality; at $\Delta Q=5$, the higher-level candidate is four words shorter in all 10 tasks. Same-subset $\Delta0$ pairs hold checklist membership fixed while varying surface realization; across the 60 pairs, absolute word-count difference is 0--5 (median 2). The target is the declared checklist, so stable response to a surface feature is cross-sensitivity only relative to this construct.

\paragraph{Exact-duplicate control interpretation.}
The tie-allowed exact-duplicate control (A0) analysis contains 120 calls per configuration: 40 empty, 40 whitespace-only, and 40 nonblank identical. Each nominal exact pair was sent in two orders. Since identical content has no meaningful order, these are repeated pipeline calls rather than two independent treatments. Empty and whitespace strata also reuse a single rendered prompt within a configuration. We report their exact point values as stress evidence and do not attach a call-binomial generalization claim.

\paragraph{Strict criterion arm.}
The original strict-prompt probe was applied only to Qwen2.5-32B on the same-subset $\Delta0$, $\Delta Q=1$, and $\Delta Q=5$ conditions. The exact-duplicate controls were not rerun under the strict prompt. Consequently, no strict-prompt \DCcore{} or \DCedge{} is reported, and the arm cannot fit a full prompt-specific psychometric curve.

\section{Analysis specification and provenance}

The fresh-item study uses \texttt{allenai/reward-bench-2} at revision \texttt{7ff08853\allowbreak b0d5686e\allowbreak79b13fda\allowbreak8677024f\allowbreak566a104a} under ODC-By. This revision identifies the dataset snapshot; the retrieved parquet was checksum-verified, with its full digest recorded in the prepared artifact. The item set, judge configurations, randomized schedule, and analysis plan were fixed before any judge calls; exact records are in the artifact. Selection yields 102 unique source rows and prompts, 17 per subset, with zero overlap against either reused-item study by row and exact or normalized prompt content. Chosen-minus-rejected absolute word difference is 0--20 (median 6.5; mean 7.26), and relative difference is at most .238 under the fixed .25 gate. The chosen answer is reused literally in the matched clone and signal arms. If a cell produces zero false preferences on 102 clones, its exact one-sided 95\% binomial upper bound is .0289. This is a fixed-item operational test rather than design-based task-population inference.

For the fresh-item signal arm, AB/BA success is averaged within each task. The subset-stratified bootstrap resamples 17 tasks within each of six subsets for 20{,}000 deterministic replicates; its 5th percentile is the one-sided lower 95\% utility bound. The randomized assignment uses six subset-balanced strata of 17 clone tasks and three strata of 34 signal tasks. Conditional randomization inference holds period outcomes fixed and reassigns realized sequence labels within these strata for 99{,}999 deterministic draws. With signal success $Y_t\in\{0,1\}$, the targeted statistics are
\[
D_C=\{(Y_{S,AB}+Y_{S,BA})-(Y_{N,AB}+Y_{N,BA})\}/2,
\]
with analogous $D_O=\{(Y_{N,BA}+Y_{S,BA})-(Y_{N,AB}+Y_{S,AB})\}/2$ and $D_I=(Y_{S,BA}-Y_{S,AB})-(Y_{N,BA}-Y_{N,AB})$, averaged over tasks. Two judge-specific $D_C$ tests form one family; both $D_O,D_I$ values for both judges form another. Clone adverse output is non-tie or invalid, and its strict-minus-neutral statistic forms a two-member family. The two-sided plus-one value is $p=(1+\sum_r\ind\{|T_r|\geq|T_{\rm obs}|\})/100{,}000$. The multiplicity families are fixed in advance; if a configuration becomes unevaluable, its associated test remains in the family with $p=1$.

The reused-item studies use a separate item set and schedule. Task-cluster intervals from the initial study remain descriptive under its deterministic schedule and post-outcome invalid-cell rule. The same-item replication fixed a Williams sequence design, which balances period and first-order carryover, before judge calls and uses analogous conditional randomization inference. Neither study is pooled with the fresh-item primary study.

\paragraph{Distinct-answer extension.}
Only the reused-item studies contain the 51-item distinct-answer \EFP{} arm. No human labels were collected for these pairs, so they contribute no estimate or test. Any future study must be specified before labeling and validate every pair without access to judge outcomes before analysis begins.

If a future construct-validated release assigns prevalence $\pi_e>0$ and loss $\lambda_e$ to the equivalence stratum, with $\pi_c+\pi_e<1$, its extended accounting map would be
\[
\begin{split}
R^{(+e)}_{J,C}={}&\pi_c\{\lambda_c\DCcore+\lambda_I\Pr(\mathrm{invalid}\mid c)\}\\
&+\pi_e\{\lambda_e\EFP+\lambda_I\Pr(\mathrm{invalid}\mid e)\}\\
&+(1-\pi_c-\pi_e)\{\lambda_W\Pr(\mathrm{wrong}\mid s)
+\lambda_T\Pr(\mathrm{tie}\mid s)+\lambda_I\Pr(\mathrm{invalid}\mid s)\}.
\end{split}
\]
This equation defines a possible future analysis; it is not used for the present results.

\paragraph{485-row extension provenance and verification.}
The 485-row protocol uses the same pinned RewardBench~2 revision as the 102-row study, with normalized-prompt exclusion and the unchanged pair-length gates. The nested set takes the first 25 tasks per subset in the frozen selection-hash order, including all 25 Ties rows. Recorded model usage was 2,477,434 prompt and 17,460 completion tokens; median client latency was .622 seconds. These are configuration-specific observations, not billing or controlled speed comparisons. The local consumer uses Ollama 0.35.0 with raw text, thinking disabled, temperature 0, seed 2026101002, output cap 2048, and context 32768. Actual prompt counts matched the precomputed vocabulary ledger for every request (163--2646 tokens). The installed tag, GGUF and executable digests identify the measured artifact; upstream checkpoint equivalence, vendor-build equivalence, and the full license chain remain unverified. The two-text serializer implements an explicit restricted template, not a validated full native-chat engine.

The frozen analysis was run once. An independent read-only verifier reconstructed all 2,910 rendered requests, raw-response joins, runtime identities, and 5,820 intent/terminal rows; a separate SQL aggregation confirmed the task and signal counts. Closed-form clone bounds and test arithmetic were independently checked. Bootstrap source, seed, replicate count, quantile rule, and reported thresholds were checked without redrawing the bootstrap. A nonterminal progress-status variable collision was retained as a monitoring defect; the final string-valued COMPLETE state, terminal receipt, and lock release were verified without changing the frozen runner or observations. Raw request/response records remain private; the checked text-free projection is prepared locally, pending final release confirmation and redistribution-license review. The earlier 612-call output-validity HOLD remains unchanged.

\section{Reproducibility, assets, compute, and LLM usage}

\paragraph{Submission-time access state.}
The accompanying offline candidate contains 1,224 sanitized outcomes and the 612-slot schedule for the primary benchmark study, with exact source verification and canonical statistical replay. It also includes the additional finite-code and context journals and a metadata-only statistical replay of the 1,224-call new-configuration benchmark audit. The prepared local-campaign layer contains 612 text-free operational records and an offline count/hash check; that check does not compute a scientific certificate. The prepared 485-row layer adds 485 text-free task records, 2,910 terminal projections, and saved analysis summaries. It verifies counts/hash-bound joins only, without raw reparse, generation, bootstrap draws, or $p$/confidence-interval recomputation. The primary replay matches the frozen 20,000-bootstrap/99,999-randomization result hash; the new-configuration replay preserves every output field except the explicitly verified public-analyzer source substitution. Eight corrupted export fixtures are rejected before statistical computation. Benchmark prompt and candidate strings, live credentials and routes, and unvalidated human annotations are excluded. Earlier primary/new-configuration layers support their stated statistical recomputation. The prepared 612-call/485-row layers support count/hash inspection only. Neither reconstructs live inference or establishes upstream-text redistribution clearance. The versioned code and releasable evaluation records are maintained at \url{https://github.com/usamihiroyasu/dark-current-v2-review}. The repository manifest identifies the accompanying files and their verification scope. The study sections and Appendices A--C specify the estimands, judge prompts, outcome schema, statistical units, validity gates, scheduling distinctions among the benchmark and code studies, randomization procedure, model provenance, and experiment-level compute used for the paper's claims.

A checksum-verified reproducibility archive has been prepared locally. Its preparation does not establish public availability. It contains machine-readable item lists and evaluation records, judge-prompt templates and parser definitions, judge-configuration records, an environment specification, verification checksums, and regeneration commands. The outgoing artifact must be checked against the final source and released with explicit code, data, and paper licensing. It does not redistribute model weights or private reasoning traces. RewardBench~2 is used under ODC-By; Llama-3.1 under the Llama~3.1 Community License; and Qwen2.5, the Qwen3.6 MLX quantization, and gpt-oss-20b under Apache~2.0, with gpt-oss usage terms retained. The legacy benchmark archive records upstream revisions or local artifact digests where retained. The new code study records its own item, schedule, source, and manifest hashes; served weight digests are unverified. These identifiers trace source artifacts; they do not prove which weight bytes a remote service used. We therefore distinguish declared artifact provenance from service-reported model identifiers.

No model is trained or fine-tuned. LLMs are the experimental judges: only the structured verdict is analyzed, and any reasoning channel is excluded. The historical controlled and Qwen2.5 evaluations used self-hosted NVIDIA RTX~4090 inference; Qwen3.6 and gpt-oss used an Apple M4~Max with 128~GB unified memory. The two fresh-item judge evaluations completed 612 calls in 281 and 855 seconds, consuming 484{,}328 and 566{,}082 API-reported tokens for Qwen2.5 and Qwen3.6, respectively. The corresponding same-item evaluations completed in 173 and 570 seconds with 220{,}264 and 294{,}427 tokens. The prepared archive records calls, tokens, wall time, and device-hours; electricity is not estimated.

The additional finite-code study used private self-hosted inference services. For its 960 calls per configuration, API-reported prompt/completion tokens are 211{,}000/5{,}760 for Qwen3.8, 259{,}392/52{,}732 for gpt-oss, and 213{,}928/187{,}648 for Nemotron, with no missing usage records. Median recorded request latencies are .889, 1.561, and 6.946 seconds, respectively; request profiles differ, so these are not controlled serving-speed comparisons. Total usage is 684{,}320 prompt and 246{,}140 completion tokens. These are experiment-level usage records and do not establish cost savings.

The context audit additionally records 135{,}504/3{,}072 prompt/completion tokens for Qwen3.8, 160{,}816/33{,}161 for gpt-oss, and 85{,}844/84{,}324 for Nemotron. The timed-out request has no usage receipt; the known sums of 382{,}164 prompt and 120{,}557 completion tokens exclude that unavailable usage. Median recorded request latencies are .899, 2.003, and 8.793 seconds. Neither the incomplete usage nor these heterogeneous request profiles supports a complete billing estimate or comparative serving claim.

The benchmark re-evaluation adds 612 calls per configuration, with 496{,}744 prompt and 3{,}679 completion tokens for Qwen3.8, and 506{,}110 prompt and 37{,}659 completion tokens for gpt-oss. All usage fields are available. Median request latencies are 2.083 and 1.870 seconds, respectively; profiles and request content preclude a controlled serving-speed comparison. These 1,224 calls are new observations on previously used task rows and are not pooled with the original benchmark responses.

\paragraph{Additional local campaign: output contract and costs.}
All 612 requests completed without transport, empty-output, or output-length failures. The frozen sole-object parser rejected 211 JSON responses wrapped in Markdown fences, and the legacy-parser check also accepted only 401. Fence removal or a format-forced retry would change the declared contract; neither is performed. Table~\ref{tab:local-output-hold} retains invalid counts with the scheduled denominators. The two criterion clone tests formed a separate two-member Holm family; they do not expand the earlier four-cell family. Neutral and strict criteria have 188/306 and 23/306 invalid responses, respectively; these are descriptive counts, not separate operational certifications. The output-validity failure identifies no semantic-quality ranking or isolated model-family, quantization, or backend effect. The public metadata do not expose response text, so readers cannot independently reparse the responses or verify the fence classification.

\begin{table}[t]
\centering\small
\caption{Invalid outputs in the additional local campaign. Each column has 102 scheduled calls per criterion. All 612 calls completed; both criteria remain HOLD under the campaign-wide operational gate. No task is added to the existing 102-row cohort.}
\label{tab:local-output-hold}
\begin{tabular}{lrrrl}
\toprule
Criterion & Clone & Signal AB & Signal BA & Scientific state\\
\midrule
Neutral & 22 & 85 & 81 & HOLD\\
Strict & 0 & 13 & 10 & HOLD\\
\bottomrule
\end{tabular}
\end{table}

The separately held local campaign used Ollama~0.35.0 with the installed \texttt{devstral-small-2:\allowbreak latest} Q4\_K\_M artifact. The local tag and blob contents were hashed, but identity with upstream checkpoint bytes and the redistribution-license chain are unverified; no weights are included. Its request profile fixed temperature zero, seed 2026101001, a 2,048-token output ceiling, and a 32,768-token context, without a forced JSON schema. All 612 responses provide usage: 485{,}592 prompt and 4{,}517 completion tokens; median client latency is 3.414 seconds. These usage records are retained despite the scientific HOLD and remain separate from engineering controls and metadata simulations.

\paragraph{Shared study assets.}
The finite task templates and executable measurement fixtures are also used in a separate code-data audit. That audit evaluates structural admission, candidate execution, and independent-suite outcomes; the present study evaluates judge decisions on matched null and signal units. Reuse of these assets is disclosed and is not counted as an independent task sample or replication. Benchmark candidate generations and judge calls remain separate measurements with their own frozen provenance.

\section{Relation to an earlier public version}

This manuscript is an expanded revision of arXiv:2606.15610v1. The controlled outcomes from that preprint are reanalyzed here. We separate null strata, use task-level units, and withdraw the claim of criterion change without resolution loss. Added evidence comprises benchmark diagnostics, the 102-row fresh-item study, randomization inference, deployment-loss analysis, two separately frozen code audits, and a new-configuration re-evaluation of the same benchmark rows. The natural-language equivalence arm remains excluded. The exhaustive-oracle study completed 2,880 calls with all six working gates passing; the context audit retains 1,344 terminal records, eight complete descriptive cells, and four held cells. The new-configuration benchmark re-evaluation adds 1,224 complete valid calls and retains all four working-gate outcomes, including the non-certified neutral gpt-oss clone bound. A further 612 completed local comparisons are retained as an output-validity HOLD, not as an additional certified configuration. The separate 485-row extension adds 2,910 valid calls and two aggregate working passes, without subgroup certification or pooling with earlier rows. These observations do not validate human equivalence or a geometric mechanism. These additions extend the same public work rather than constituting an independent duplicate publication.

\section{Exploratory Ineligible-Reference Context Audit}
\label{sec:v7}

Preferential-ranking consistency, including independence from irrelevant alternatives, has already been studied \citep{zhao2024inconsistency}. We ask a narrower question: does a functional-equivalence judgment remain stable when an ineligible reference is displayed, and how does useful discrimination change alongside it? A reference can supply information despite being ineligible to win, so the operational context comparison does not equate every change with a formal independence violation.

\paragraph{Separately frozen design.}
Before inspecting the preceding code study's scientific outcomes, we fixed 32 items: parameter indices 0, 3, 6, and 9 from each of its eight families. Each configuration receives 512 planned calls (32 tasks $\times$ two criteria $\times$ two arms $\times$ two orders $\times$ two contexts). Reference C is appended after A/B and cannot win. On equivalence pairs C is verified faulty; on signal pairs C is the alternative verified correct implementation. The A/B contents, specification, rubric, and output schema stay identical between absent and present contexts. Both contexts receive the same added instruction to compare A/B only. This makes the absent baseline contemporaneous within this audit, but changes the prompt relative to the 96-item code study; a between-study difference is not the context effect. A fixed shuffled schedule (seed 2026100807) is reused across configurations. This is not a test of all positions or full triad rankings.

\paragraph{Endpoints and incomplete execution.}
A task is context-adverse if its canonical winner changes between contexts in either order or any of its four responses is invalid. Success is a tie on equivalence pairs and a correct-reference choice on signal pairs. We average AB/BA success within task, then report present-minus-absent differences and descriptive 95\% intervals from 20,000 bootstraps within eight four-item families. No additional confirmatory test or multiplicity family is created. Each configuration must complete all 512 calls with reconciled served identity and at most 1\% invalid output before its outcome cells are admitted.

The audit stops with 1,344/1,536 terminal records and no ambiguous intent. Qwen3.8 and gpt-oss each complete 512/512 with zero invalid output. Nemotron completes 320/512, with one length-truncated empty response and one timed-out request. The last request has a 2.09-second recorded latency at the end of approximately 90 minutes; the frozen runner bounds request timeout by the remaining time window. The preserved terminal state is \texttt{PARTIAL\_HOLD\_TRANSPORT\_OR\_MODEL}. The timeout has no served-model identifier, so identity reconciliation is incomplete; a named replacement model was not observed. All four Nemotron cells remain HOLD. Their partial task groups do not supply success rates or intervals, and no request is retried.

\begin{table}[!htbp]
\centering\small
\setlength{\tabcolsep}{3pt}
\begin{tabular}{lllrrrl}
\toprule
Configuration & Criterion & Arm & Adverse/32 & $U_a$ & $U_p$ & Difference [95\% interval] \\
\midrule
Qwen3.8 & neutral & Equiv. & 7 & 1.0000 & 0.8594 & $-0.1406\;[-0.1875,-0.0938]$ \\
Qwen3.8 & neutral & Signal & 4 & 0.9219 & 0.9844 & $0.0625\;[0.0312,0.0938]$ \\
Qwen3.8 & strict & Equiv. & 1 & 1.0000 & 0.9688 & $-0.0312\;[-0.0938,0.0000]$ \\
Qwen3.8 & strict & Signal & 4 & 0.9219 & 0.9844 & $0.0625\;[0.0312,0.0938]$ \\
gpt-oss & neutral & Equiv. & 0 & 1.0000 & 1.0000 & $0.0000\;[0.0000,0.0000]$ \\
gpt-oss & neutral & Signal & 0 & 1.0000 & 1.0000 & $0.0000\;[0.0000,0.0000]$ \\
gpt-oss & strict & Equiv. & 0 & 1.0000 & 1.0000 & $0.0000\;[0.0000,0.0000]$ \\
gpt-oss & strict & Signal & 0 & 1.0000 & 1.0000 & $0.0000\;[0.0000,0.0000]$ \\
\bottomrule
\end{tabular}
\medskip
\begin{tabular}{llrrrr}
\toprule
Held configuration & Criterion/arm & Calls/128 & Full tasks/32 & Invalid absent/present & Status \\
\midrule
Nemotron & neutral/equiv. & 86 & 6 & 0/0 & HOLD \\
Nemotron & neutral/signal & 78 & 4 & 1/0 & HOLD \\
Nemotron & strict/equiv. & 78 & 5 & 0/0 & HOLD \\
Nemotron & strict/signal & 78 & 4 & 1/0 & HOLD \\
\bottomrule
\end{tabular}
\caption{All twelve context-audit cells, including four held cells. Each admitted cell has 128/128 calls, 32 complete tasks, and zero invalid output. Adverse counts a canonical context change in either order or invalidity among four responses. $U_a$ and $U_p$ are task-averaged success in absent and present contexts. Paired intervals are descriptive, from the frozen 20,000-replicate bootstrap. The four Nemotron cells report only coverage and invalidity; partial task groups do not yield performance estimates.}
\label{tab:v7-results}
\end{table}

\paragraph{Observed context contrasts.}
For Qwen3.8's neutral equivalence arm, adverse tasks are 7/32 and tie success falls from 1.000 to .859375, a paired difference of $-.140625$ with descriptive interval $[-.1875,-.09375]$. The strict equivalence arm has 1/32 adverse tasks and difference $-.03125$ with interval $[-.09375,0]$. In both signal arms, adverse tasks are 4/32, while success rises from .921875 to .984375, difference $.0625$ with interval $[.03125,.09375]$. Thus context changes are not uniformly harmful, and improved signal utility does not establish preservation of the equivalence null. For gpt-oss, all four cells have 0/32 adverse tasks and success one in both contexts. Its zero-width resampling intervals express the observed constant outcomes, not a general invariance guarantee.

These are exploratory fixed-schedule observations on constructed code. Serving stability, order/time effects, unseen domains, and stochastic repetition remain unverified; the intervals are not simultaneous population guarantees. We do not pool the 96-item code-study or benchmark observations, or claim context instability as a new general phenomenon.

\end{document}